\documentclass[conference]{IEEEtran}
\IEEEoverridecommandlockouts
\usepackage{cite}
\usepackage{amsmath,amssymb,amsfonts}

\usepackage{graphicx}
\usepackage{textcomp}
\usepackage{xcolor}

\graphicspath{ {./images/} }
\usepackage{algorithm}
\usepackage{algpseudocode}
\usepackage{subcaption}

\def\BibTeX{{\rm B\kern-.05em{\sc i\kern-.025em b}\kern-.08em
    T\kern-.1667em\lower.7ex\hbox{E}\kern-.125emX}}
\begin{document}

\title{REFORMER: A ChatGPT-Driven Data Synthesis Framework Elevating Text-to-SQL Models}

\author{\IEEEauthorblockN{Shenyang Liu}
\IEEEauthorblockA{\textit{Department of Computer Science} \\
\textit{University of Central Florida}\\
Orlando, USA \\
shenyang.liu@ucf.edu}
\and
\IEEEauthorblockN{Saleh Almohaimeed}
\IEEEauthorblockA{\textit{Department of Computer Science} \\
\textit{University of Central Florida}\\
Orlando, USA \\
sa247216@ucf.edu}
\and
\IEEEauthorblockN{Liqiang Wang}
\IEEEauthorblockA{\textit{Department of Computer Science} \\
\textit{University of Central Florida}\\
Orlando, USA \\
liqiang.wang@ucf.edu}
}

\maketitle
\renewcommand{\thefootnote}{}
\footnotetext{\makebox[\columnwidth]{1946-0759/24/\$31.00~\copyright2024 IEEE \hfill} \hspace{\columnsep}\makebox[\columnwidth]{ }}

\begin{abstract}
The existing Text-to-SQL models suffer from a shortage of training data, inhibiting their ability to fully facilitate the applications of SQL queries in new domains. To address this challenge, various data synthesis techniques have been employed to generate more diverse and higher quality data. In this paper, we propose REFORMER, a framework that leverages ChatGPT's prowess without the need for additional training, to facilitate the synthesis of (question, SQL query) pairs tailored to new domains. Our data augmentation approach is based on a ``retrieve-and-edit" method, where we generate new questions by filling masked question using explanation of SQL queries with the help of ChatGPT. Furthermore, we demonstrate that cycle consistency remains a valuable method of validation when applied appropriately. Our experimental results show that REFORMER consistently outperforms previous data augmentation methods. To further investigate the power of ChatGPT and create a general data augmentation method, we also generate the new data by paraphrasing the question in the dataset and by paraphrasing the description of a new SQL query that is generated by ChatGPT as well. Our results affirm that paraphrasing questions generated by ChatGPT help augment the original data.
\end{abstract}


\section{Introduction}
For Text-to-SQL tasks, popular training datasets like WikiSQL \cite{zhong2017seq2sql} and Spider \cite{yu2018spider} are often too small, limiting model generalization \cite{hazoom2021text} \cite{suhr2020exploring}. This makes models struggle in new domains. To address this, data synthesis is crucial. Unlike past SQL-to-Text research that relied on zero-shot methods \cite{yang2021hierarchical} \cite{zhao2022importance}, we believe using one-shot learning with examples can enhance performance. Previous data synthesis methods also lacked query diversity, reducing model effectiveness \cite{zhong2020grounded} \cite{wu2021data} \cite{wang2021learning} \cite{yang2021hierarchical}.

We introduce REFORMER, a framework leveraging ChatGPT for data synthesis. By retrieving similar (question, SQL query) pairs and masking the questions, we use ChatGPT to generate new queries without fine-tuning. A novel cycle-consistency validation approach compares the similarity between ChatGPT-generated questions and SQL query explanations, improving stability over past methods \cite{awasthi2022diverse}. Additionally, we explore two data augmentation techniques: paraphrasing existing questions and creating new SQL queries using schema templates.

Our contributions include: (1) to the best of our knowledge, we are the first one to create a SQL-to-Text framework REFORMER using ChatGPT without fine-tuning, (2) we introduce question-query-question cycle consistency validation, (3) we show REFORMER's improvement over other data augmentation methods, and (4) demonstrate effective data generation with two paraphrasing techniques.

\section{Related Work}

\subsection{SQL-to-Text Task}
The models for SQL-to-Text task include those built upon the Gated Recurrent Unit (GRU) \cite{guo2018question}, LSTM networks \cite{zhong2020grounded} \cite{wu2021data}, and advanced models like BART \cite{shi2021learning} and T5 \cite{yang2021hierarchical} \cite{zhao2022importance}. In specific cases, some models, as demonstrated by \cite{zhong2020grounded}, leverage BERT \cite{devlin2018bert} as an encoder to enhance overall performance. The SQL queries used in training these models can come from a variety of sources. Some models employ hand-crafted templates, while others leverage grammars extracted from existing SQL queries \cite{wu2021data} or draw from a corpus of SQL queries obtained through web scraping or crawling open-source code repositories \cite{shi2021learning}. The authors of \cite{zhang2023sciencebenchmark} and us independently came up with the same question-query-question cycle consistency as in our paper concurrently, but they train a GPT-3 model and do not use the method for data augmentation purpose but for validation in text-to-SQL task.

\subsection{Large Language Models and Prompt Engineering For SQL-to-Text}

It is noteworthy that, to date, there has not been many research \cite{yang2021hierarchical}\cite{zhao2022importance}\cite{zhang2023sciencebenchmark} considering the utilization of state-of-the-art Large Language Models (LLMs) for SQL-to-text tasks. Our work, therefore, can be viewed as pioneering in this regard. By harnessing the capabilities of LLMs, we aim to inspire and open up new avenues for further investigations into data augmentations using these new models.

Prompt-based learning directly uses the model without training or fine-tuning to get the answer for the questions. Prompt engineering usually uses manually designed templates \cite{cui2021template} \cite{lester2021power} or optimization techniques \cite{petroni2019language} to construct more effective prompts for the model.

\section{Data synthesis with REFORMER}

\begin{figure}[h]
\centering
\includegraphics[width=\columnwidth]{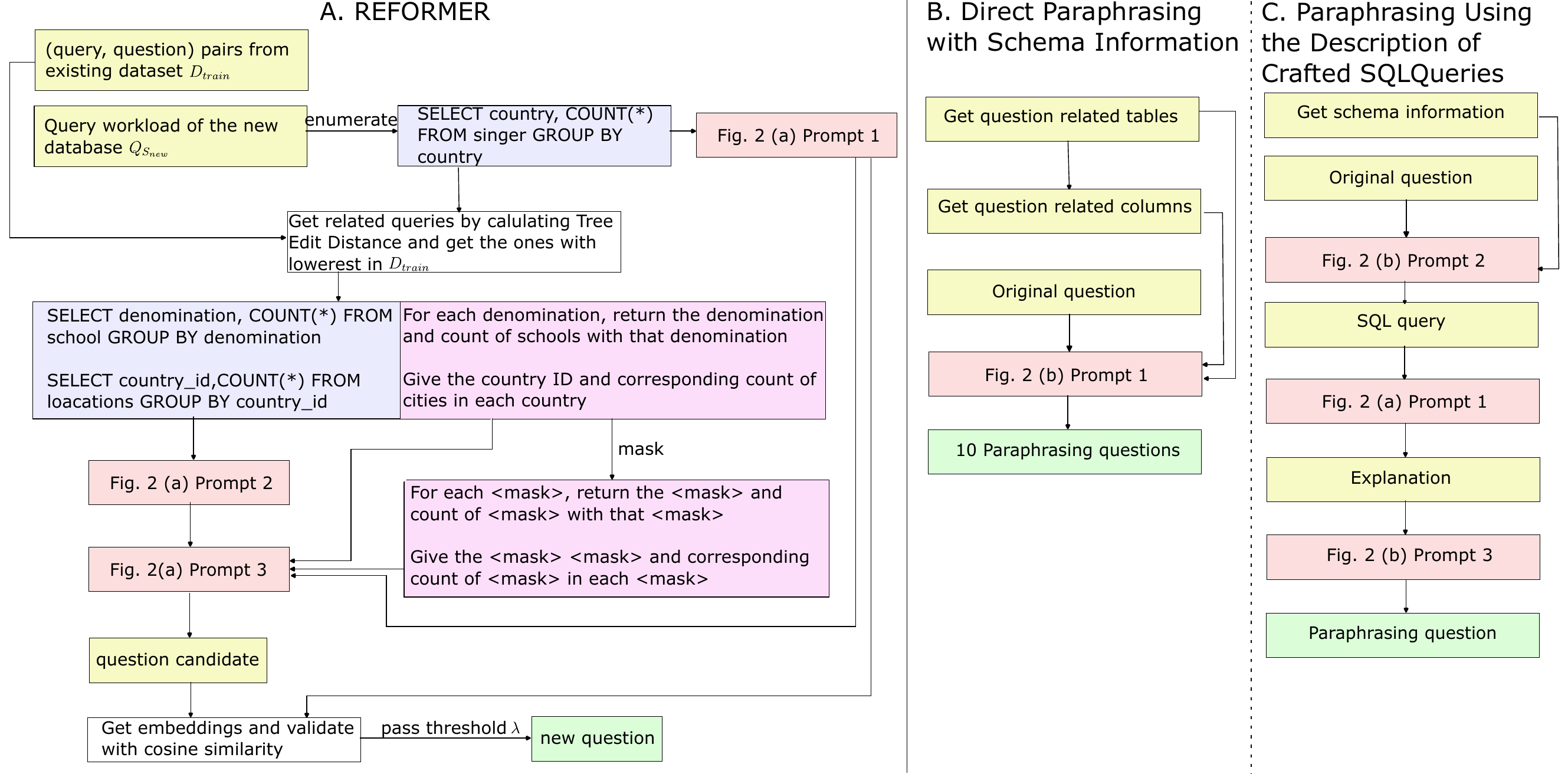}
\caption{Framework for REFORMER and Paraphrasing-based Data Sythesis methods}
\label{fig:prompts_flowchart_simplified}
\end{figure}

In this study, we focus on data synthesis for Text-to-SQL parsers by leveraging the power of GPT-based models. Directly training a SQL-to-Text model on existing data and using manually crafted templates for generating new SQL queries has limitations that the results may fail to accurately capture the underlying data distribution present in the original dataset due to the lack of guided question generation. To address this issue, we adopt the ``retrieve-and-edit" approach introduced by \cite{awasthi2022diverse} and improve the approach to avoid training models. Our approach is summarized in Algorithm~\ref{alg:REFORMER} and the whole framework is shown in Figure~\ref{fig:prompts_flowchart_simplified}A. More detailed explanations for the algorithm are provided from Section 3.1 to Section 3.3.

\begin{algorithm}
\caption{Data Augmentation with REFORMER}\label{alg:REFORMER}
\hspace*{\algorithmicindent} \textbf{Input} $Q_{S_{new}}, M, D_{train}$
\begin{algorithmic}
\State $D_{new} \gets \emptyset$
\For{\texttt{$q \gets$ enumerate($Q_{S_{new}}$)}}
\State \texttt{${q_{sim}, u_{sim}} \gets$ \textbf{GetRelatedQueries}($q$, $D_{train}$)}
\State \texttt{${u_{temp}} \gets$ \textbf{MaskSchemaTokens}({$q_{sim}, u_{sim}$})}
\State \texttt{$expl1 \gets$ \textbf{GetExplanation}($q$)}
\State \texttt{$u_{result} \gets$ \textbf{FillTemplate}($u_{temp}$, $expl1$)}
\State \texttt{$expl2 \gets$ \textbf{GetExplanation}($q$)}
\State \texttt{$D_{new} \gets D_{new} \cup$ \textbf{Validate}($expl2$, $u_{result}$)}
\EndFor
\State \texttt{$M_{new} \gets$ \textbf{Fine-tune}($M, D_{new}$)}
\end{algorithmic}
\end{algorithm}

\subsection{Related SQL queries and question templates preparation}

The SQL queries we use within a new schema $S_{new}$ can be achieved by manually creating or using a grammar-based SQL generator \cite{zhong2020grounded} \cite{wang2021learning} using random parts of the schema information, denoted as $Q_{S_{new}}$. We consider a Text-to-SQL parser model $M$ trained by $D_{train}$ to be enhanced by our proposed method.

{\bf GetRelatedQueries:} Given a query $q$, we aim to identify a SQL query with high structural similarity (denoted by $q_{sim}$) to $q$ and its corresponding question (or utterance denoted by $u_{sim}$). Our goal is to derive question templates that ensure fidelity to the training data distribution. To retrieve SQL queries exhibiting high structural similarity to a given query $q$, we adopt the method in \cite{awasthi2022diverse} that utilizes the tree-edit-distance \cite{pawlik2015efficient}. This tree-edit-distance measures the similarity between two relational algebra trees constructed from SQL queries. The smaller the distance, the higher the structural similarity between the queries. This approach does not consider schema-specific information to assess structural similarity, allowing us to focus on the query structure itself. We adopt the hyperparameters in \cite{awasthi2022diverse} so that we only retain queries with a distance of less than 0.1 from the given query $q$, ensuring that they exhibit substantial structural resemblance. 

{\bf MaskSchemaTokens:} The question templates play an important role in guiding our data synthesis process. These templates are derived from existing questions but with schema-specific information masked out. The purpose of these templates is to guide the generation of questions while ensuring that the resulting questions conform to the data distribution observed in the training set. We adopt the frequency-based method in \cite{awasthi2022diverse} to determine which words should be included in the question templates and which should be masked. Specifically, we retain all words that appear in more than 50\% of the schemas, as these words are indicative of common patterns and characteristics in the training data. Any words that fall below this threshold are masked and replaced with a special token, denoted as ``MASK". Consecutive masked words are represented by a single ``MASK" token to maintain conciseness and readability. For a question $u$, we get template $u_{temp}$ after masking based on the explanation above. This preparation of related SQL queries and question templates serves as the foundation for our data synthesis methodology, enabling the generation of high-quality synthetic data to enhance the performance of Text-to-SQL parsers.

\subsection{New Question Generation}
{\bf GetExplanation} and {\bf FillTemplate}: We generate new questions for Text-to-SQL parsers by leveraging ChatGPT to fill masked templates with one-sentence query explanations. Using Prompt 1 in Figure~\ref{fig:prompts}, ChatGPT extracts a concise explanation $expl1$ from a given SQL query $q \in Q_{S_{new}}$, resulting in question $u_{result}$. Since the Spider dataset typically has succinct, one-sentence questions, we match this format.

We improve upon \cite{awasthi2022diverse} by avoiding intermediate representations, which often include unnecessary terms (e.g., "belongs to" for tables, "the number of" for COUNT(*)) that negatively affect ChatGPT’s performance. Instead, we instruct ChatGPT to explain the SQL in a single sentence without using table names, yielding more accurate question generation. We employ in-context learning \cite{brown2020language}, providing original (question, SQL query) pairs to guide ChatGPT, allowing it to generate questions aligned with the desired structure. This method helps create a new dataset $D_{new}$ for fine-tuning existing Text-to-SQL parsers.

\subsection{Result validation}
{\bf GetExplanation} and {\bf Validate}: To ensure the consistency of the generated SQL queries and the associated questions, we came up with a validation approach that leverages the embeddings produced by ChatGPT. Rather than training a dedicated model to validate the alignment between SQL queries and newly generated questions, we employ embeddings generated from ChatGPT. Specifically, we compute the cosine similarity between the embeddings derived from the one-sentence explanation of the SQL query and the generated question. To increase the diversity of the questions, we generate a new one-sentence query explanation $expl2$ based on the corresponding Prompt 2 in Figure~\ref{fig:prompts} for the cosine similarity rather than use the one to generate the result question. This method unveils the relationship between SQL queries and natural language questions, effectively establishing what can be termed as a ``question-query-question cycle" consistency. Notably, our approach differs from previous methods \cite{zhong2020grounded} that directly train binary classifiers with (question, SQL query) pairs. The drawback of such direct training is the absence of specific structures tailored for SQL queries. Unlike models designed specifically for SQL queries \cite{wang2019rat} \cite{chen2021shadowgnn}, where the classifiers might struggle to elucidate the nuanced relationship between queries and questions, our method addresses this limitation.
We agree with the concerns raised by \cite{awasthi2022diverse} about the usage of ``query-question-query" cycle-consistency, which get a query to generate a question, and utilize a Text-to-SQL model to generate a query and then compare the similarity between the two queries. However, we demonstrate that the concept of cycle consistency can be valuable when applied appropriately, as evident in our methodology. To ensure the coherence and consistency of the generated questions, we employ a threshold for cosine similarity, similar to previous work \cite{awasthi2022diverse}. This threshold serves as a filter, allowing us to identify and retain only the questions that exhibit a satisfactory degree of consistency with their corresponding SQL queries. The questions that pass this filtering process, along with their associated queries, are then utilized for {\bf fine-tuning} existing Text-to-SQL model $M$ with a new dataset $D_{new}$ to get a better model $M_{new}$.

\section{Paraphrasing-Based Data Synthesis Method Using ChatGPT}
To fully leverage the natural language processing capabilities of ChatGPT, we explore two general data augmentation methods that utilize the paraphrase with schema information to generate new data without considering a specific domain. 
The method is shown in Fig. \ref{fig:prompts_flowchart_simplified}B.

\subsection{Direct Paraphrasing with Schema Information}
Traditionally, paraphrasing methods focused solely on the sentences themselves. \cite{dopierre2021protaugment} \cite{dai2023chataug} However, for datasets like Spider, where questions are associated with specific schema information, we propose an enhanced approach. Beyond the question text, we utilize ChatGPT in conjunction with the LangChain library to extract related tables from the schema for a given question. Subsequently, we retrieve all relevant column names for these tables. The paraphrasing process involves incorporating the question, related table names, and column names, ensuring that the generated data is closely aligned with the underlying database structure. Figure~\ref{fig:prompts_paraphrase_full} Prompt 1 illustrates the prompt used in this approach. To maintain the result quality, the same cycle-consistency validation method is employed here.

\subsection{Paraphrasing Using the Description of Crafted SQL Queries}
\begin{figure}[h]
\centering
\begin{subfigure}{\columnwidth}
\includegraphics[width = \columnwidth]{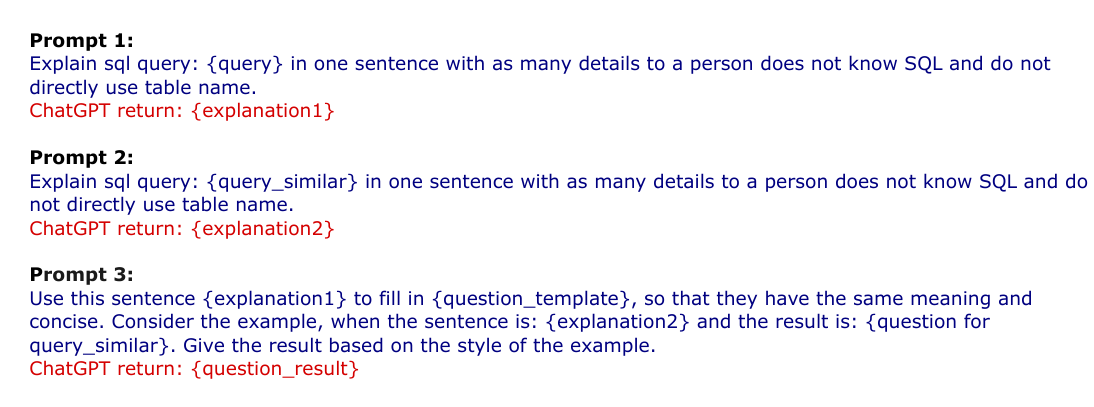}
\centering
\caption{Prompts for REFORMER}
\label{fig:prompts}
\end{subfigure}
\begin{subfigure}{\columnwidth}
\includegraphics[width = \columnwidth]{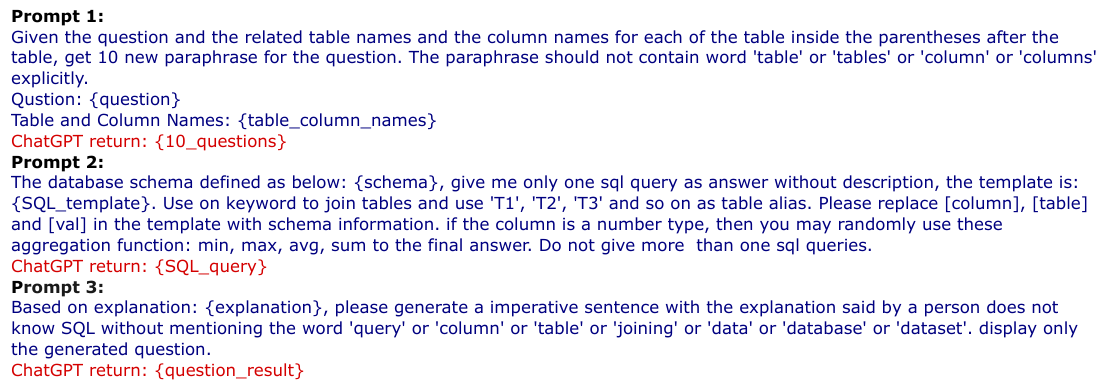}
\centering
\caption{Prompts for Paraphrasing-based Method}
\label{fig:prompts_paraphrase_full}
\end{subfigure}
\caption{Prompts for data generation, where ``{query}", ``{query\_similar}", ``{explanation 1/2}", ``{question\_result}", ``{question\_template}" etc. denote the corresponding placeholders, which will be replaced by real queries or explanations when handing SQL query and question pairs.}
\end{figure}

Different from the direct paraphrasing method, we introduce a strategy based on crafting SQL query templates. Starting with basic queries containing select, from, where, group by, order by, and limit clauses, we progress to generating complex queries with in, union, except, and intersect clauses. Following the creation of these templates, ChatGPT is employed to fill in the blanks with tables, columns and values, guided by the related schema. The synthesized SQL queries are then executed to validate their correctness, and only those without errors are retained. For each SQL query, we use ChatGPT to generate a concise one-sentence description. New questions are subsequently generated by paraphrasing these descriptions. The prompt used for this method is detailed in Figure~\ref{fig:prompts_paraphrase_full} Prompt 2 and Prompt 3.

\section{Experiment Setting}
\subsection{Datasets}
For our experiments, we use the Spider dataset \cite{yu2018spider}, which includes 7,000 (question, SQL query) pairs in the training set and 1,034 pairs in the dev set. Following prior work \cite{awasthi2022diverse}, we categorize database schemas by domain, such as music, pets, university records, and country records. We also replace constants in 70\% of the SQL queries with other values from the same columns, as in \cite{awasthi2022diverse}. For paraphrasing-based methods, we use the Spider dataset directly without modifications, as these methods are not domain-specific.

\subsection{Text-to-SQL Model}
In keeping with previous studies \cite{awasthi2022diverse}, we employ the SmBop model \cite{rubin2020smbop} as our baseline Text-to-SQL model. The SmBop model utilizes the RoBERTA \cite{liu2019roberta} model as its backbone and connects it with four Relation-Aware Attention (RAT) \cite{wang2019rat} layers. The dev set in the Spider dataset is modified to ensure that it does not include any data associated with the evaluation categories.

\subsection{SQL-to-Text Model}
Our approach capitalizes on the capabilities of ChatGPT \cite{chatgpt} for the generation of text-based outputs. As detailed in Section 3.2, we utilize ChatGPT to fill the masked templates, and a sample prompt is illustrated in Fig.~\ref{fig:prompts}. For paraphrasing-based method, we use ChatGPT to generate related tables, SQL queries and paraphrases, and a sample prompt is shown in Fig.~\ref{fig:prompts_paraphrase_full}.

\subsection{Validation Module}
The validation module (outlined in Section 3.3) computes the cosine similarity between the GPT embeddings (text-embedding-ada-002) derived from the one-sentence explanation of the SQL query and the generated question. A threshold value, denoted as $\lambda$, is set to 0.85. This threshold aids in filtering out inconsistent results, enhancing the overall quality of our synthesized data. Beyond the threshold for similarity, we keep only the best 5 results to maintain the high quality of the generated data used for fine-tuning. The method of ``Direct Paraphrasing with Schema Information" uses a larger threshold $\lambda$ for the similarity to decrease the size of the dataset. We tried different values of $\lambda$ (such as 0.9, 0.93, 0.95) in our experiments.

\subsection{Baselines}
For REFOMER framework, we incorporate ReFill \cite{awasthi2022diverse} as the baseline model in our experiments. ReFill has demonstrated superior performance compared to several previous methods, such as L2S \cite{wang2021learning}, GAZP \cite{zhong2020grounded}, and SNOWBALL \cite{shu2021logic}. 
For our paraphrasing-based data synthesis method, we compare with SmBop model\cite{rubin2020smbop}, T5 model+Picard\cite{scholak2021picard} and T5 model+Picard+synthesized data\cite{zhao2022importance}.

\subsection{Evaluation Metrics}
In line with the Spider dataset's evaluation methodology \cite{yu2018spider}, we evaluate the performance of our models using Exact Set Match (EM) and Execution Accuracy (EX). EM disregards database-related values and compares the result query with the gold query word by word. On the other hand, EX executes both the result query and the gold query via a SQL engine and measures their concordance.

\section{Experimental Results and analysis}
\begin{table}
\centering
\caption{
Results for fine-tuning the SmBop parser on (question, SQL query) pairs generated using REFILL and REFORMER}\label{result1}
\resizebox{\columnwidth}{!}{
\begin{tabular}{llllll}
\hline
\textbf{} & \textbf{Category1 (Music)} & \textbf{Category2 (Pets)} & \textbf{Category3 (University)} & \textbf{Category4 (Country)} & \textbf{Average}\\
\hline
Method & EM/EX & EM/EX & EM/EX & EM/EX & EM/EX\\
\hline
SmBop\cite{rubin2020smbop} &  87.8/88.7 & 63.7/65.3 & 69.5/69.5 & 44.2/35.8 & 66.3/65.0\\
REFILL\cite{awasthi2022diverse}(In Paper) &  88.7/87.0 & \textbf{69.7}/\textbf{73.8} & 73.2/70.1 & 55.8/45.0 & \textbf{71.8}/68.9\\
REFILL\cite{awasthi2022diverse}(Experiment) & 85.2/86.1 & 56.5/58.1 & 67.7/67.7 & 56.7/46.7 & 66.4/64.7\\
Ours (validation only) & 87.8/88.7 & 56.5/57.3 & 72.6/73.2 & 54.2/46.7 & 67.9/66.8\\
Ours (data augmentation only) & 87.8/88.7 & 59.7/62.9 & 72.6/73.8 & 56.7/47.5 & 69.2/68.5\\
Ours & \textbf{89.6}/\textbf{89.6} & 58.9/62.1 & \textbf{73.8}/\textbf{74.4} & \textbf{57.5}/\textbf{50.8} & 70.0/\textbf{69.4}\\
\hline
\end{tabular}
}
\end{table}

\subsection{Overall Evaluation}
Our overall evaluation, detailed in Table~\ref{result1}, demonstrates the superior performance of our method compared to ReFill across different categories. Notably, our approach outperforms ReFill in all categories in our own experiment (denoted by ``Experiment"), while surpassing the claimed performance in most categories (except for category 2) in \cite{awasthi2022diverse} (denoted by ``In Paper"). Further scrutiny of the data in Category 2 highlights potential reasons for performance deterioration with data augmentation methods. On average, we achieve an increase in Exact Set Match (EM) by 3.6\% and Execution Accuracy (EX) by 4.7\% when compared to the REFILL framework (Experiment). We do not beat EM for REFILL (In Paper) but beat EX by 0.5\%.

\subsection{Generated Question Evaluation}
To have a fair comparison, we only compare our method with REFILL(Experiment). Table~\ref{result1} shows the performance of using the SQL-to-Text module of the REFORMER framework independently (data augmentation only) and then using the REFILL filtering module. The results reveal improvements in EM by 2.8\% and EX by 3.8\%. This analysis underscores the effectiveness of our data augmentation approach, which contributes to the overall performance enhancement.

\subsection{Cycle Consistency Method Evaluation}
In Table~\ref{result1}, we examine the performance when utilizing the validation module of the REFORMER framework exclusively (validation only). The results illustrate improvements in EM by 1.5\% and EX by 2.1\%. Notably, comparing with the performance improvement by our approach with only data augmentation (denoted by ``Ours (data augmentation only)"), our approach using only cycle consistency validation method (denoted by ``Ours (validation only)") is less effective.

\subsection{Error Analysis for REFORMER}
Our analysis reveals that these errors disproportionately impact the data augmentation in Category 2, ultimately yielding suboptimal results (see Table~\ref{result1}). Notably, some of the generated templates retain information from the original questions should be excluded, which may result in inaccuracies. However, for brevity, we focus solely on errors attributable to other causes.

The REFILL method in our experiment introduces multiple challenges. Specifically, we observe the questions generated in Category 2 in our experiment are less complex than the data synthesized in \cite{awasthi2022diverse}. This discrepancy in data distribution adversely affects the performance of the model. Moreover, certain errors manifest more frequently in our generated data than in the data synthesized by \cite{awasthi2022diverse}.

Our method REFORMER also has its own limitations in terms of data diversity and the reliance on one-shot settings. While data diversity is generally a desirable aspect of data augmentation, it can inadvertently disrupt existing models due to unseen novel expressions during training. The one-shot setting, where substantial emphasis is placed on only one example, restricts the flexibility to incorporate additional information into the question template.

\begin{figure}[h]
\centering
\includegraphics[width = \columnwidth]{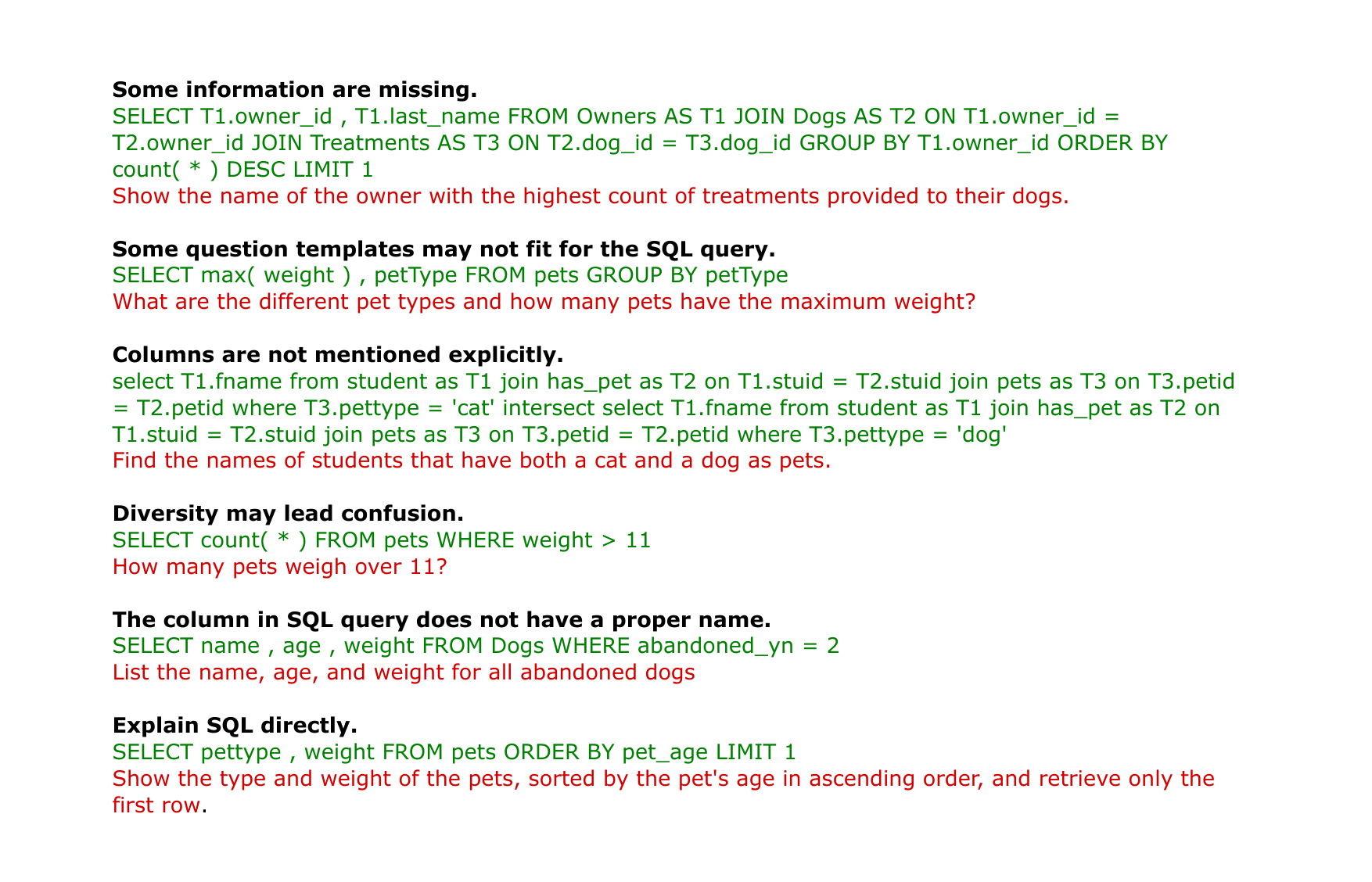}
\caption{Errors from REFORMER}
\label{fig:REFORMER_errors}
\end{figure}

In Figure~\ref{fig:REFORMER_errors}, we comprehensively present the errors identified by our experiment on REFORMER. The specific errors examined are as follows:

\textbf{``Some Information Are Missing" Error:}
This occurs when the SQL query specifies the last name, but the question only asks for "name." Consistent rules for specifying "first name" or "last name" could help, though expecting perfect precision is challenging.

\textbf{``Question Template Mismatch" Error:}
A mismatch arises when the SQL query retrieves the maximum weight for pets, but the question template introduces irrelevant phrases like "how many," which doesn't align with the intent of the SQL query.

\textbf{``Columns Not Explicitly Mentioned" Error:}
This error happens when important columns, like student IDs, are needed in the SQL query but are not mentioned in the corresponding question.

\textbf{``Diversity Leads to Confusion" Error:}
ChatGPT's varied phrasing, like using "weigh" instead of "weight," can confuse the model when it expects specific column names.

\textbf{``Improper Column Names" Error:}
SQL queries may use abbreviations (e.g., "yn" for year number), and while ChatGPT can interpret common ones, unfamiliar abbreviations may cause failures.

\textbf{``Explain SQL Directly" Error:}
Generated questions sometimes directly reflect SQL descriptions in an unnatural, non-conversational tone, highlighting the need for more human-like question generation.

\subsection{Quality and Diversity of Generated Questions}
To gauge the quality and diversity of the generated questions, we employ two metrics. The BLEU score \cite{papineni2002bleu} of the set $S(q)$ measures the quality of the synthesized data for a query $q$ when compared to its corresponding gold query. Concurrently, the SelfBLEU metric \cite{zhu2018texygen} calculates the average BLEU score among $S(q)$, providing a measure of the diversity of the synthesized data. Higher BLEU scores denote higher quality, while lower SelfBLEU scores indicate greater diversity.

In the study of \cite{awasthi2022diverse}, the generation of hypotheses for each query involves producing ten candidates through beam sampling \cite{fan2018hierarchical}. The selection of the final hypothesis is determined by the highest BLEU score, serving as our final result for evaluation. In our comparative analysis, we also consider L2S \cite{wang2021learning}, GAZP \cite{zhong2020grounded}, and SNOWBALL \cite{shu2021logic}. It's essential to note that our approach lacks a dedicated training phase, which makes the direct comparisons with other methods unfair. Instead, we leverage all possible templates to generate hypotheses, selecting the one yielding the highest BLEU score. 

The results in Table~\ref{result2} indicate that our method does not surpass REFILL in terms of BLEU and SelfBLEU scores. However, it outperforms other methods in BLEU. Despite the lower score on BLEU and higher score on SelfBLEU, our method demonstrates superiority over REFILL in terms of Exact Match (EM) and Execution Accuracy (EX) across most categories.

\subsection{Result Evaluation for Direct Paraphrasing with Schema Information}
The findings (see Table~\ref{result3}) show the performance of our method highly depends on the size of the augmented data.

When considering values of $\lambda$ set at 0.93 and 0.95, corresponding to augmented data sizes of 21,781 and 8,939, respectively, the performance closely aligns with the baseline SmBop model, exhibiting marginal improvement of less than 1\%. This observation suggests that, within certain data augmentation thresholds, the Direct Paraphrasing method does not significantly outperform the SmBop baseline.

However, a distinct shift is discerned when $\lambda$ is set to 0.9, resulting in an augmented data size of 43,847. In this scenario, the results exhibit marked improvements, with Exact Match (EM) showing a commendable increase of 2.8\% and the Exact F1 (EX) metric registering a corresponding rise of 2.7\%. These enhancements bring the EM metric close to the performance of T5-3B with PICARD \cite{scholak2021picard} and surpass the performance of T5-3B with Syn Data \cite{zhao2022importance}.

\subsection{Issues for Paraphrasing using the Explanation of Crafted SQL Queries}

The SmBop model excels due to its use of parsing trees for both input and output validation of SQL queries. However, its custom parser is limited to specific cases, leading to errors when parsing more complex SQL templates in our new dataset. To address this, we switched to RASAT(-small), a model based on T5-small, due to resource constraints preventing us from testing on T5-3B.

Our results showed that using the Explanation of Crafted SQL Queries for paraphrasing significantly decreased EM and EX scores. The errors mainly occurred with hard-level SQL queries, so we designed even more complex templates than those in the Spider dataset. However, the lack of ground truth for the generated data contributed to performance issues.

\begin{table}[!htb]
    \caption{Results for BLEU and Self-BLEU score from different methods
    }\label{result2}
    \resizebox{\columnwidth}{!}{
    \begin{tabular}{lll}
    \hline
    \textbf{Method} & \textbf{BLEU $\uparrow$ (Quality)} & \textbf{100-SelfBLEU $\uparrow$ (Diversity)} \\
    \hline
    Gold-Ref & 100 & 68.8 \\
    \hline
    L2S & 38.0 & 2.2\\
    GAZP & 38.8 & 2.0\\
    SNOWBALL & 40.2 & 2.8\\
    REFILL & 48.6 & 33.8\\
    Ours & 43.2 & 41.0\\
    \hline
    \end{tabular}
    }
\end{table}
\begin{table}
    \caption{
    Results for Paraphrasing-based method
    }\label{result3}
    \resizebox{\columnwidth}{!}{
    \begin{tabular}{lll}
    \hline
    \textbf{} & Training Set Size & EM/EX\\
    \textbf{Base(SmBop\cite{rubin2020smbop})} & 7000 & 72.1/72.3\\
    \textbf{With Schema Info ($\lambda$=0.9)} & 7000+43847 & 74.9/75.0\\
    \textbf{With Schema Info ($\lambda$=0.93)} & 7000+21871 & 72.6/72.3\\
    \textbf{With Schema Info ($\lambda$=0.95)} & 7000+8939 & 72.1/71.5\\
    \textbf{With Crafted SQL} & 7000+12173 & 25.7/33.7\\
    \hline
    \textbf{T5-3B\cite{scholak2021picard}} & 7000 & 71.5/74.4\\
    \textbf{T5-3B + PICARD\cite{scholak2021picard}} & 7000 & 75.5/79.3\\
    \textbf{T5-3B + Syn Data\cite{zhao2022importance}} & 7000+21851 & 74.5/78.6\\
    \textbf{T5-3B + PICARD + Syn Data\cite{zhao2022importance}} & 7000+21851 & 76.1/81.4\\
    \end{tabular}
    }
\end{table}

\section{Conclusion and Limitation}
In this study, we introduced REFORMER for generating (question, SQL query) pairs tailored to new domains using a "retrieve-and-edit" approach. By filling question templates with information from queries and leveraging ChatGPT with carefully crafted prompts, REFORMER outperforms previous methods without extra training. Our experiments highlight the effectiveness of cycle-consistency for validation. Additionally, we explored two paraphrasing-based data synthesis methods, demonstrating ChatGPT’s ability to generate high-quality data. While we only focused on ChatGPT with spider dataset and manually crafted prompts, future research should explore automatic prompt generation and test other LLMs and Text-to-SQL datasets.
\footnotetext{Citation: S. Liu, S. Almohaimeed and L. Wang, "REFORMER: A ChatGPT-Driven Data Synthesis Framework Elevating Text-to-SQL Models," 2024 International Conference on Machine Learning and Applications (ICMLA), Miami, FL, USA, 2024, pp. 828-833, doi: 10.1109/ICMLA61862.2024.00119. IEEE Xplore link: https://ieeexplore.ieee.org/abstract/document/10903391}
\bibliographystyle{IEEEtran}
\bibliography{custom}

\begin{thebibliography}{10}
\providecommand{\url}[1]{#1}
\csname url@samestyle\endcsname
\providecommand{\newblock}{\relax}
\providecommand{\bibinfo}[2]{#2}
\providecommand{\BIBentrySTDinterwordspacing}{\spaceskip=0pt\relax}
\providecommand{\BIBentryALTinterwordstretchfactor}{4}
\providecommand{\BIBentryALTinterwordspacing}{\spaceskip=\fontdimen2\font plus
\BIBentryALTinterwordstretchfactor\fontdimen3\font minus \fontdimen4\font\relax}
\providecommand{\BIBforeignlanguage}[2]{{%
\expandafter\ifx\csname l@#1\endcsname\relax
\typeout{** WARNING: IEEEtran.bst: No hyphenation pattern has been}%
\typeout{** loaded for the language `#1'. Using the pattern for}%
\typeout{** the default language instead.}%
\else
\language=\csname l@#1\endcsname
\fi
#2}}
\providecommand{\BIBdecl}{\relax}
\BIBdecl

\bibitem{zhong2017seq2sql}
V.~Zhong, C.~Xiong, and R.~Socher, ``Seq2sql: Generating structured queries from natural language using reinforcement learning,'' \emph{arXiv:1709.00103}, 2017.

\bibitem{yu2018spider}
T.~Yu, R.~Zhang, K.~Yang, M.~Yasunaga, D.~Wang, Z.~Li, J.~Ma, I.~Li, Q.~Yao, S.~Roman \emph{et~al.}, ``Spider: A large-scale human-labeled dataset for complex and cross-domain semantic parsing and text-to-sql task,'' \emph{arXiv:1809.08887}, 2018.

\bibitem{hazoom2021text}
M.~Hazoom, V.~Malik, and B.~Bogin, ``Text-to-sql in the wild: a naturally-occurring dataset based on stack exchange data,'' \emph{arXiv:2106.05006}, 2021.

\bibitem{suhr2020exploring}
A.~Suhr, M.-W. Chang, P.~Shaw, and K.~Lee, ``Exploring unexplored generalization challenges for cross-database semantic parsing,'' in \emph{Proceedings of the 58th Annual Meeting of the Association for Computational Linguistics}, 2020, pp. 8372--8388.

\bibitem{yang2021hierarchical}
W.~Yang, P.~Xu, and Y.~Cao, ``Hierarchical neural data synthesis for semantic parsing,'' \emph{arXiv:2112.02212}, 2021.

\bibitem{zhao2022importance}
Y.~Zhao, J.~Jiang, Y.~Hu, W.~Lan, H.~Zhu, A.~Chauhan, A.~Li, L.~Pan, J.~Wang, C.-W. Hang \emph{et~al.}, ``Importance of synthesizing high-quality data for text-to-sql parsing,'' \emph{arXiv:2212.08785}, 2022.

\bibitem{zhong2020grounded}
V.~Zhong, M.~Lewis, S.~I. Wang, and L.~Zettlemoyer, ``Grounded adaptation for zero-shot executable semantic parsing,'' \emph{arXiv:2009.07396}, 2020.

\bibitem{wu2021data}
K.~Wu, L.~Wang, Z.~Li, A.~Zhang, X.~Xiao, H.~Wu, M.~Zhang, and H.~Wang, ``Data augmentation with hierarchical sql-to-question generation for cross-domain text-to-sql parsing,'' \emph{arXiv:2103.02227}, 2021.

\bibitem{wang2021learning}
B.~Wang, W.~Yin, X.~V. Lin, and C.~Xiong, ``Learning to synthesize data for semantic parsing,'' \emph{arXiv:2104.05827}, 2021.

\bibitem{awasthi2022diverse}
A.~Awasthi, A.~Sathe, and S.~Sarawagi, ``Diverse parallel data synthesis for cross-database adaptation of text-to-sql parsers,'' \emph{arXiv:2210.16613}, 2022.

\bibitem{guo2018question}
D.~Guo, Y.~Sun, D.~Tang, N.~Duan, J.~Yin, H.~Chi, J.~Cao, P.~Chen, and M.~Zhou, ``Question generation from sql queries improves neural semantic parsing,'' \emph{arXiv:1808.06304}, 2018.

\bibitem{shi2021learning}
P.~Shi, P.~Ng, Z.~Wang, H.~Zhu, A.~H. Li, J.~Wang, C.~N. dos Santos, and B.~Xiang, ``Learning contextual representations for semantic parsing with generation-augmented pre-training,'' in \emph{Proceedings of the AAAI Conference on Artificial Intelligence}, vol.~35, no.~15, 2021, pp. 13\,806--13\,814.

\bibitem{devlin2018bert}
J.~Devlin, M.-W. Chang, K.~Lee, and K.~Toutanova, ``Bert: Pre-training of deep bidirectional transformers for language understanding,'' \emph{arXiv preprint arXiv:1810.04805}, 2018.

\bibitem{zhang2023sciencebenchmark}
Y.~Zhang, J.~Deriu, G.~Katsogiannis-Meimarakis, C.~Kosten, G.~Koutrika, and K.~Stockinger, ``Sciencebenchmark: A complex real-world benchmark for evaluating natural language to sql systems,'' \emph{arXiv preprint arXiv:2306.04743}, 2023.

\bibitem{cui2021template}
L.~Cui, Y.~Wu, J.~Liu, S.~Yang, and Y.~Zhang, ``Template-based named entity recognition using bart,'' \emph{arXiv:2106.01760}, 2021.

\bibitem{lester2021power}
B.~Lester, R.~Al-Rfou, and N.~Constant, ``The power of scale for parameter-efficient prompt tuning,'' \emph{arXiv:2104.08691}, 2021.

\bibitem{petroni2019language}
F.~Petroni, T.~Rockt{\"a}schel, P.~Lewis, A.~Bakhtin, Y.~Wu, A.~H. Miller, and S.~Riedel, ``Language models as knowledge bases?'' \emph{arXiv:1909.01066}, 2019.

\bibitem{pawlik2015efficient}
M.~Pawlik and N.~Augsten, ``Efficient computation of the tree edit distance,'' \emph{ACM Transactions on Database Systems (TODS)}, vol.~40, no.~1, pp. 1--40, 2015.

\bibitem{brown2020language}
T.~Brown, B.~Mann, N.~Ryder, M.~Subbiah, J.~D. Kaplan, P.~Dhariwal, A.~Neelakantan, P.~Shyam, G.~Sastry, A.~Askell \emph{et~al.}, ``Language models are few-shot learners,'' \emph{Advances in neural information processing systems}, vol.~33, pp. 1877--1901, 2020.

\bibitem{wang2019rat}
B.~Wang, R.~Shin, X.~Liu, O.~Polozov, and M.~Richardson, ``Rat-sql: Relation-aware schema encoding and linking for text-to-sql parsers,'' \emph{arXiv:1911.04942}, 2019.

\bibitem{chen2021shadowgnn}
Z.~Chen, L.~Chen, Y.~Zhao, R.~Cao, Z.~Xu, S.~Zhu, and K.~Yu, ``Shadowgnn: Graph projection neural network for text-to-sql parser,'' \emph{arXiv:2104.04689}, 2021.

\bibitem{dopierre2021protaugment}
T.~Dopierre, C.~Gravier, and W.~Logerais, ``Protaugment: Intent detection meta-learning through unsupervised diverse paraphrasing,'' in \emph{ACL-IJCNLP}.\hskip 1em plus 0.5em minus 0.4em\relax ACL, 2021, pp. 2454--2466.

\bibitem{dai2023chataug}
H.~Dai, Z.~Liu, W.~Liao, X.~Huang, Z.~Wu, L.~Zhao, W.~Liu, N.~Liu, S.~Li, D.~Zhu \emph{et~al.}, ``Chataug: Leveraging chatgpt for text data augmentation,'' \emph{arXiv:2302.13007}, 2023.

\bibitem{rubin2020smbop}
O.~Rubin and J.~Berant, ``Smbop: Semi-autoregressive bottom-up semantic parsing,'' \emph{arXiv:2010.12412}, 2020.

\bibitem{liu2019roberta}
Y.~Liu, M.~Ott, N.~Goyal, J.~Du, M.~Joshi, D.~Chen, O.~Levy, M.~Lewis, L.~Zettlemoyer, and V.~Stoyanov, ``Roberta: A robustly optimized bert pretraining approach,'' \emph{arXiv preprint arXiv:1907.11692}, 2019.

\bibitem{chatgpt}
\BIBentryALTinterwordspacing
OpenAI. (2022) Chatgpt (sep 25 version). [Online]. Available: \url{https://chat.openai.com/chat}
\BIBentrySTDinterwordspacing

\bibitem{shu2021logic}
C.~Shu, Y.~Zhang, X.~Dong, P.~Shi, T.~Yu, and R.~Zhang, ``Logic-consistency text generation from semantic parses,'' \emph{arXiv:2108.00577}, 2021.

\bibitem{scholak2021picard}
T.~Scholak, N.~Schucher, and D.~Bahdanau, ``Picard: Parsing incrementally for constrained auto-regressive decoding from language models,'' \emph{arXiv:2109.05093}, 2021.

\bibitem{papineni2002bleu}
K.~Papineni, S.~Roukos, T.~Ward, and W.-J. Zhu, ``Bleu: a method for automatic evaluation of machine translation,'' in \emph{ACL}, 2002, pp. 311--318.

\bibitem{zhu2018texygen}
Y.~Zhu, S.~Lu, L.~Zheng, J.~Guo, W.~Zhang, J.~Wang, and Y.~Yu, ``Texygen: A benchmarking platform for text generation models,'' in \emph{The 41st international ACM SIGIR conference on research \& development in information retrieval}, 2018, pp. 1097--1100.

\bibitem{fan2018hierarchical}
A.~Fan, M.~Lewis, and Y.~Dauphin, ``Hierarchical neural story generation,'' \emph{arXiv:1805.04833}, 2018.

\end{thebibliography}

\end{document}